\documentclass{article}
\usepackage{spconf,amsmath,graphicx,hyperref}
\usepackage{amssymb,algorithm,booktabs,multirow,adjustbox,subcaption}

\title{CT-Merging: Consensus Directions and Task-Specific Scaling for LoRA Adapter Merging}
\name{Keumseo Ryum \qquad Joonhyuk Kang\thanks{ This work was partly supported by the Institute of Information \& Communications Technology Planning \& Evaluation (IITP)-ITRC (Information Technology Research Center) grant funded by the Korea government (MSIT) (IITP-2026-RS-2020-II201787, contribution rate: 50\%) and (IITP-2026-RS-2023-00259991, contribution rate: 50\%).}}
\address{KAIST \\
Daejeon, Republic of Korea\\ keumseo@kaist.ac.kr \quad jkang@kaist.ac.kr}
\begin{document}
%
\maketitle
\begin{abstract}
LoRA merging methods increasingly operate on the low-rank structure of task updates, yet how the common subspace is estimated and how coefficients are assigned after recomposition are rarely compared directly. We propose CT-Merging, which estimates common directions from averaged task subspace projectors and assigns a separate residual scale to each task. Projector averaging selects directions supported across task subspaces without weighting them by singular magnitude, while task-specific scaling removes component-wise magnitude variation and preserves scale differences across tasks. On the released KnOTS CLIP adapters, CT-Merging achieves the best average and worst-task normalized accuracy on both backbones, improving over the strongest baseline by up to 2.56 and 6.65 points, respectively. On the DC-Merge adapter benchmark, it achieves the best average normalized accuracy in eight of nine backbone and task-count settings. Ablations show that projector averaging outperforms summed-update SVD and that task-specific scaling improves worst-task accuracy over global isotropic scaling.
\end{abstract}
%



\section{Introduction}
Large pretrained vision and vision-language models are increasingly specialized to downstream tasks through parameter-efficient fine-tuning, as full fine-tuning is costly in memory and storage. Low-rank adaptation (LoRA) \cite{hu2022lora} represents each task update using two low-rank matrices, allowing a pretrained model to support many tasks through lightweight adapters. As the number of downstream tasks grows, however, maintaining a separate adapter for every task increases storage and requires selecting the appropriate adapter at inference time.

Model merging provides a way to combine independently trained task updates into a single model without access to joint training data. Early methods average fine-tuned checkpoints or combine task vectors derived from a common pretrained model \cite{ilharco2023editing}, while later methods reduce conflicts among task updates through pruning, sign resolution, or learned merging weights \cite{yadav2023tiesmerging}. These methods were largely developed for full fine-tuned updates, and their performance does not directly transfer to independently trained LoRA adapters, whose low-rank updates can exhibit substantially weaker alignment \cite{stoica2025knots}.

Recent LoRA merging methods therefore operate directly on the low-rank structure of task updates. Some align LoRA updates in shared representations with lower dimension \cite{stoica2025knots,panariello2025accurate}, while others modify singular coefficients or improve directional consistency before merging \cite{zeng2025parameter,zhang2026dcmerge}. Although these methods specify their own subspaces and coefficient rules, direct comparisons of how the common subspace is estimated and how coefficients are assigned after recomposition remain limited.

Moreover, merging performance is commonly summarized by average accuracy or average normalized accuracy across tasks. Evaluating based on average accuracy can mask whether performance gains are broadly distributed or concentrated in a subset of tasks, and does not reveal the lowest task performance of the merged model. This distinction matters when several task adapters are combined into a single model, since a high aggregate score can coexist with a substantial degradation on one task. We therefore report worst-task normalized accuracy together with the average and examine how coefficient assignment affects both metrics.


To address these challenges, we propose CT-Merging (Consensus directions and Task-specific scaling), a LoRA-aware merging method that estimates common directions from averaged task subspace projectors and assigns a separate scale to the residual directions of each task. Projector-based consensus selects directions supported across task subspaces, while task-specific scaling removes component-wise magnitude variation without collapsing residual scale differences across tasks. Experiments on released DC-Merge and KnOTS checkpoints show the best average normalized accuracy in most settings and strong worst-task performance, with controlled ablations supporting both components of the method. Our contributions are as follows.

\begin{itemize}
\item We estimate common directions from averaged task subspace projectors and show that this construction outperforms common bases obtained from summed task updates on both evaluated backbones.
\item We assign a separate residual scale to each task while removing component-wise coefficient variation, and show that this choice improves worst-task normalized accuracy over global isotropic scaling.
\item We evaluate our method on released DC-Merge and KnOTS checkpoints across multiple ViT backbones and task counts, reporting both average and worst-task normalized accuracy against recent LoRA merging methods.
\end{itemize}

\begin{figure}[t]
\centering
\includegraphics[width=\linewidth]{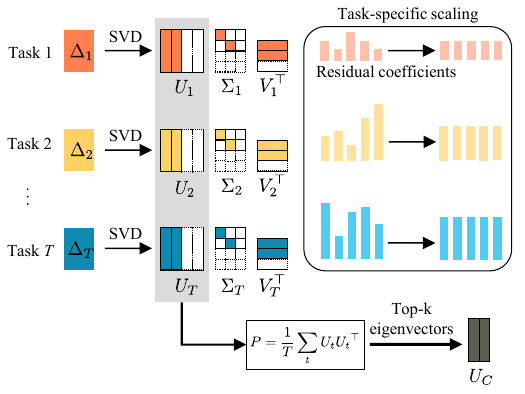}
\caption{
Two design components of CT-Merging. Averaged task projectors determine the common subspace, while task-specific scaling assigns equal coefficient magnitudes to the residual directions of each task and preserves scale differences across tasks.
}
\label{fig:overview}
\end{figure}

\section{Related Work}
\label{sec:related}
Model merging methods combine independently trained updates through parameter arithmetic or conflict resolution \cite{ilharco2023editing,yadav2023tiesmerging}. Spectral approaches instead analyze task matrices through singular directions and subspaces, including isotropic merging and orthogonal recomposition \cite{marczak2025no,gargiulo2024tsv}. Recent parameter efficient methods further address the alignment, representation, and singular coefficients of low-rank adapters \cite{stoica2025knots,zeng2025parameter,zhang2026dcmerge,panariello2025accurate}. CT-Merging follows the data free setting and focuses on common subspace estimation and residual coefficient assignment after SVD based recomposition.



\section{Methodology}
\label{sec:method}

\subsection{Setup}
We consider model merging by conducting arithmetic operations on task updates obtained by subtracting the pretrained weights from the weights fine-tuned on task $t$ \cite{ilharco2023editing}. Let \(W_0\) and \(W_t\) denote the pretrained and task-specific weights, respectively. The task update is \(\Delta_t=W_t-W_0\), which is represented by the LoRA factors as

\begin{equation}
\Delta_t = B_tA_t.
\end{equation}
We denote its retained SVD factors by
\(
\Delta_t \approx U_tS_tV_t^\top,
\)
where we use the retained left singular vectors to estimate the common subspace from which $k$ common directions are selected. We separately retain the leading $r$ singular triplets for residual construction.
\begin{table}[t]
\centering
\caption{
Average and worst-task normalized accuracy on the CLIP adapters provided by KnOTS. Best results are in bold.
}
\label{tab:knots-results-Vitb}
\setlength{\tabcolsep}{5pt}
\begin{tabular}{lcc|cc}
\toprule
& \multicolumn{2}{c|}{ViT-B/32}
& \multicolumn{2}{c}{ViT-L/14} \\
\cmidrule(lr){2-3}
\cmidrule(lr){4-5}
Method & Avg & Worst & Avg & Worst \\
\midrule
TA
& 63.61 & 41.28
& 73.29 & 55.31 \\

TIES
& 63.78 & 37.25
& 75.50 & 60.86 \\

KnOTS-TIES
& 67.32 & 47.95
& 78.68 & 65.03 \\

Iso-CTS
& 74.59 & 44.07
& 85.88 & 70.68 \\

Core Space
& 76.36 & 50.59
& 86.00 & 71.38 \\

DC-Merge
& 77.48 & 52.66
& 85.76 & 74.15 \\
\midrule
{CT-Merging}
& \textbf{80.04} & \textbf{59.31}
& \textbf{87.27} & \textbf{76.33} \\
\bottomrule
\end{tabular}
\end{table}
\begin{table*}[t]
\centering
\small
\caption{
Average normalized accuracy on the DC-Merge adapter benchmark.
Best value in each column is bold.
}
\label{tab:task_scaling_dcmerge}
\setlength{\tabcolsep}{4.5pt}
\begin{tabular*}{\textwidth}{@{\extracolsep{\fill}}lccc|ccc|ccc}
\toprule
& \multicolumn{3}{c|}{ViT-B/16}
& \multicolumn{3}{c|}{ViT-B/32}
& \multicolumn{3}{c}{ViT-L/14} \\
\cmidrule(lr){2-4}
\cmidrule(lr){5-7}
\cmidrule(lr){8-10}
Method
& 8 tasks & 12 tasks & 16 tasks
& 8 tasks & 12 tasks & 16 tasks
& 8 tasks & 12 tasks & 16 tasks \\
\midrule
TA
& 64.83 & 71.50 & 71.89
& 61.42 & 69.08 & 69.54
& 74.61 & 78.81 & 77.43 \\
TIES
& 63.47 & 68.56 & 64.98
& 61.14 & 67.92 & 65.60
& 75.48 & 78.93 & 74.97 \\
KnOTS-TIES
& 64.52 & 70.94 & 70.99
& 56.51 & 65.58 & 67.02
& 72.79 & 78.47 & 76.95 \\
Iso-CTS
& 77.19 & 77.92 & 77.30
& 72.38 & 74.08 & 73.61
& 88.53 & 86.38 & 83.36 \\
Core Space
& 77.97 & 78.05 & 77.27
& 71.42 & 74.10 & 73.60
& 88.45 & 84.33 & 81.74 \\

DC-Merge
& 78.47 & 79.90 & {78.63}
& \textbf{72.86} & 75.55 & {74.70}
& 89.80 & 89.39 & {85.90} \\
CT-Merging
& \textbf{80.51} & \textbf{80.85} & \textbf{79.22}
& 72.55 & \textbf{76.00} & \textbf{75.02}
& \textbf{90.19} & \textbf{89.69} & \textbf{86.22} \\
\bottomrule
\end{tabular*}
\vspace{2pt}
\end{table*}

\subsection{Consensus and Residual Directions}

The common subspace should reflect overlap among the retained task subspaces without further weighting their directions by singular magnitude. We average the corresponding subspace projectors as
\begin{equation}
P=
\frac{1}{T}
\sum_{t=1}^{T}
U_tU_t^\top,
\label{eq:avproj}
\end{equation}
and take its leading $k$ eigenvectors
\begin{equation}
U_c=\operatorname{TopEig}_k(P).
\end{equation}
A basis of this eigenspace maximizes the aggregate projection onto the retained task subspaces
\begin{equation}
U_c \in
\arg\max_{Q^\top Q=I_k}
\sum_{t=1}^{T}
\|Q^\top U_t\|_F^2.
\end{equation}

The common basis \(U_c\) contains only left directions. To obtain the corresponding right directions, we project each task update onto \(U_c\) and compute the SVD of their average,

\begin{equation}
\frac{1}{T}\sum_{t=1}^{T}U_c^\top\Delta_t
=
P_c\Sigma_cV_c^\top,
\end{equation}
where $P_c$ and $V_c$ contain the left and right singular vectors, respectively.
We then update \(U_c\leftarrow U_cP_c\) and use \(V_c\) as the corresponding right directions. This rotation preserves the common subspace and its projector \(U_cU_c^\top\).

For each task, we project the leading $r$ left singular vectors onto the orthogonal complement of the common subspace. For singular triplet $(u_{t,j},s_{t,j},v_{t,j})$, we define
\begin{equation}
z_{t,j}
=
(I-U_cU_c^\top)u_{t,j},
\qquad
a_{t,j}=\|z_{t,j}\|_2,
\end{equation}
and
\begin{equation}
\bar u_{t,j}
=
\frac{z_{t,j}}{a_{t,j}},
\qquad
\bar s_{t,j}
=
a_{t,j}s_{t,j}.
\end{equation}
The coefficient adjustment preserves the projected residual update
\begin{equation}
R_t
=
\sum_{j=1}^{r}
\bar s_{t,j}\bar u_{t,j}v_{t,j}^\top
=
(I-U_cU_c^\top)
\sum_{j=1}^{r}
s_{t,j}u_{t,j}v_{t,j}^\top .
\end{equation}

Collecting the residual directions as
$
\bar U_t=[\bar u_{t,1},\ldots,\bar u_{t,r}]
$
and
$
V_t^{(r)}=[v_{t,1},\ldots,v_{t,r}],
$
we concatenate
\begin{equation}
U=[U_c\mid\bar U_1\mid\cdots\mid\bar U_T],
\qquad
V=[V_c\mid V_1^{(r)}\mid\cdots\mid V_T^{(r)}].
\end{equation}
Following TSV \cite{gargiulo2024tsv}, we orthogonalize both direction matrices using SVD based whitening. For \(X=L_X\Sigma_XR_X^\top,\)
we define \( \mathcal{W}(X)=L_XR_X^\top\)
and set \(\widetilde U=\mathcal{W}(U)\)
and \(\widetilde V=\mathcal{W}(V).\)

\subsection{Task-Specific Scaling}

After residual construction and orthogonalization, CT-Merging assigns coefficients to the recomposed directions. Retaining separate component magnitudes preserves fine grained variation from the task decompositions, while a single global magnitude also removes scale differences across tasks. We therefore preserve the projected residual energy of each task and distribute it uniformly across its $r$ residual directions.

The projected residual energy of task $t$ is
\begin{equation}
E_t
=
\|R_t\|_F^2
=
\sum_{j=1}^{r}
\bar s_{t,j}^{\,2}.
\end{equation}
We assign a common magnitude $\rho_t$ to its $r$ residual directions while preserving this energy,
\begin{equation}
\rho_t
=
\sqrt{\frac{E_t}{r}},
\qquad
D_t=\rho_t I_r.
\end{equation}
Thus,
$
\|D_t\|_F^2=E_t,
$
while different tasks retain different residual scales. The scalar form is also unchanged under paired orthogonal reparameterizations of the residual coordinates.

The common directions use the same energy-preserving rule as in,
\begin{equation}
E_c=\|\Sigma_c\|_F^2,
\qquad
\rho_c=\sqrt{\frac{E_c}{k}}.
\end{equation}
Following the ordering of the concatenated common and residual coordinates, we define
\begin{equation}
D=
\operatorname{diag}
\left(
\rho_c I_k,
\rho_1 I_r,
\ldots,
\rho_T I_r
\right).
\end{equation}
The merged task update is
\begin{equation}
\Delta_{\mathrm{merge}}
=
\gamma
\widetilde U
D
\widetilde V^\top,
\end{equation}
where $\gamma$ controls the overall merge magnitude. The final merged weights are then computed as \( W_0+\Delta_{\mathrm{merge}}.\) Figure \ref{fig:overview} visualizes the mechanism of our method.

\section{Experiments}
\label{sec:exp}
\subsection{Settings}
We evaluate CT-Merging on released CLIP LoRA checkpoints from KnOTS \cite{stoica2025knots} and DC-Merge \cite{zhang2026dcmerge}, covering ViT-B/16, ViT-B/32, and ViT-L/14 across 8-, 12-, and 16-task settings. All adapters use LoRA rank 16, and we retain the corresponding 16 singular directions when constructing the task subspaces. Since the merged update is added to W0, the recomposition rank k + Tr does not change the size of the merged model. We therefore fix (k, r) for each backbone across task counts, using (8, 15) for the ViT-B backbones and (16, 14) for ViT-L/14.

The 8-task vision benchmark from \cite{stoica2025knots} consists of Cars \cite{cars}, DTD \cite{dtd}, EuroSAT \cite{eurosat}, GTSRB \cite{gtrsb}, MNIST \cite{mnist}, RESISC45 \cite{resisc45}, SUN397 \cite{sun397}, and SVHN \cite{svhn}, while the 12-task setting adds CIFAR100 \cite{cifar10}, Flowers102 \cite{oxfordflowers}, OxfordIIITPet \cite{OxfordIIITPet} and STL10 \cite{stl10} to the 8-task vision benchmark, and the 16-task experiment adds four datasets to the 12-task configuration: FER2013 \cite{fer2013}, CIFAR10 \cite{cifar10}, FashionMNIST \cite{fashionmnist}, and RenderedSST2 \cite{renderedsst2}.

We compare our method against Task Arithmetic (TA) \cite{ilharco2023editing}, TIES Merging \cite{yadav2023tiesmerging}, KnOTS-TIES  \cite{stoica2025knots}, Iso-CTS \cite{marczak2025no}, Core Space merging \cite{panariello2025accurate}, and DC-Merge \cite{zhang2026dcmerge}, where we use hyperparameters reported in their original papers. For each method, we select \(\gamma\) from \(\gamma\in\{0.1,0.2,\ldots,1.0,1.3,1.5\}\) by maximizing the average normalized accuracy and report both the average normalized accuracy and worst-task normalized accuracy at the selected value. The normalized accuracy for task $t$ is defined as the accuracy of the merged model on task $t$ divided by the accuracy of the individual model on task $t$.

\subsection{Main Results}

Table \ref{tab:knots-results-Vitb} presents CT-Merging on the ViT-B/32 and ViT-L/14 checkpoints provided by KnOTS \cite{stoica2025knots} on the 8-task vision merging benchmark, where we observe that our method achieves the highest average normalized accuracy compared to baselines, improving over the best baseline by up to 2.56 points. We also highlight that our method results in the highest \textit{worst-task normalized accuracy}, suggesting that CT-Merging provides not only overall performance gains but also improved performance on relatively weak tasks.

Table~\ref{tab:task_scaling_dcmerge} reports average normalized accuracy on the DC-Merge checkpoints. CT-Merging achieves the highest score in eight of nine settings, with the 8-task setting on ViT-B/32 as the only exception. Together with the results on the KnOTS checkpoints, these findings show that CT-Merging maintains consistently strong performance across different checkpoint sources, backbones, and task counts.



\begin{table}[t]
\centering
\caption{
Consensus source ablation on the DC-Merge CLIP ViT eight-task setting. Only the source of the common basis is varied.
}
\label{tab:consensus-ablation}
\setlength{\tabcolsep}{4pt}
\begin{tabular}{lcc|cc}
\toprule
& \multicolumn{2}{c|}{ViT-B/32} & \multicolumn{2}{c}{ViT-L/14} \\
\cmidrule(lr){2-3}
\cmidrule(lr){4-5}
Source & Avg & Worst & Avg & Worst \\
\midrule
Avg-projector & \textbf{72.55} & \textbf{51.78} & \textbf{90.19} & \textbf{84.20} \\
Random &  70.42 & 43.28 & 88.00 & 76.28 \\
Svd-sum & 68.86 & 43.42 & 87.30 & 75.73 \\
\bottomrule
\end{tabular}
\end{table}
\begin{table}[t]
\centering
\small
\caption{
Coefficient allocation on fixed recomposition directions.
Component-wise coefficients are rescaled to match the residual energy of each task.
}
\label{tab:coeff-ablation}
\setlength{\tabcolsep}{4pt}
\begin{tabular}{lcc|cc}
\toprule
& \multicolumn{2}{c|}{ViT-B/32}
& \multicolumn{2}{c}{ViT-L/14} \\
\cmidrule(lr){2-3}
\cmidrule(lr){4-5}
Coefficient rule & Avg & Worst & Avg & Worst \\
\midrule
Component-wise
& 73.61 & 52.71
& 83.60 & 72.76 \\

Global isotropic
& 79.18 & 48.52
& \textbf{89.26} & 71.19 \\

Task-specific
& \textbf{80.04} & \textbf{59.31}
& 87.27 & \textbf{76.33} \\
\bottomrule
\end{tabular}
\end{table}
\subsection{Ablation studies}


Table~\ref{tab:consensus-ablation} isolates the source used to construct the common subspace while keeping the remaining merging procedure fixed. \textit{Avg-projector} is based on Eq. (\ref{eq:avproj}), \textit{Svd-sum} uses the leading left singular vectors of the summed task update, and \textit{Random} uses an orthonormal basis of the same rank. \textit{Avg-projector} achieves the highest average and worst-task normalized accuracy on both backbones. Projector averaging outperforms the summed-update SVD on both metrics and backbones, supporting the use of subspace overlap to estimate common directions. The random basis also fails to match \textit{Avg-projector}, confirming that the result depends on which common directions are selected.

Table~\ref{tab:coeff-ablation} compares three coefficient rules with fixed recomposition directions on KnOTS adapter checkpoints on the 8-task benchmark. \textit{Component-wise} retains individual residual coefficients after matching the residual energy of each task. \textit{Global isotropic} uses one magnitude across all residual directions. \textit{Task-specific} removes component-wise variation while retaining a separate scale for each task. \textit{Task-specific} outperforms the component-wise rule in both average and worst-task accuracy on the two backbones, supporting assigning equal coefficient magnitudes to each task's residual directions in the evaluated settings. Compared with global isotropic scaling, separate task scales yield substantially higher worst-task normalized accuracy on both backbones, while average normalized accuracy increases on ViT-B/32 and decreases on ViT-L/14. 

\section{Conclusion}

We propose CT-Merging, a LoRA-adapter merging method that keeps shared and task residual directions while replacing component-wise magnitudes with a separate scale per task. Compared to state-of-the-art methods, CT-Merging achieves the best average normalized accuracy in 10 of 11 settings and the best worst-task normalized accuracy on the KnOTS checkpoints. Ablation studies support both the consensus construction and the task-specific coefficient assignment.

\vfill\pagebreak

\bibliographystyle{IEEEbib}
\bibliography{strings,refs}

\end{document}